\documentclass[
]{ceurart}
\usepackage{tikz}
\usetikzlibrary{shapes, arrows}
\usepackage{graphicx}
\usepackage{listings}
\begin{document}

\copyrightyear{2023}
\copyrightclause{Copyright for this paper by its authors.
  Use permitted under Creative Commons License Attribution 4.0
  International (CC BY 4.0).}

\conference{CLEF 2023: Conference and Labs of the Evaluation Forum, September 18–21, 2023, Thessaloniki, Greece}

\title{MASON-NLP at eRisk 2023: Deep Learning-Based Detection of Depression Symptoms from Social Media Texts}

\title[mode=sub]{Notebook for the MasonNLP Lab at CLEF 2023}

\author[1]{Fardin Ahsan Sakib}[%
email=fsakib@gmu.edu,
]
\fnmark[1]
\address[1]{George Mason University, Fairfax, VA}

\author[1]{Ahnaf Atef Choudhury}[%
email=achoudh9@gmu.edu,
]
\fnmark[1]

\author[1]{Ozlem Uzuner}[%
email=ouzuner@gmu.edu
]

\fntext[1]{These authors contributed equally.}

\begin{abstract}
Depression is a mental health disorder that has a profound impact on people's lives. Recent research suggests that signs of depression can be detected in the way individuals communicate, both through spoken words and written texts. In particular, social media posts are a rich and convenient text source that we may examine for depressive symptoms. The Beck Depression Inventory (BDI) Questionnaire, which is frequently used to gauge the severity of depression, is one instrument that can aid in this study. We can narrow our study to only those symptoms since each BDI question is linked to a particular depressive symptom. It's important to remember that not everyone with depression exhibits all symptoms at once, but rather a combination of them. Therefore, it is extremely useful to be able to determine if a sentence or a piece of user-generated content is pertinent to a certain condition. With this in mind, the eRisk 2023 Task 1 was designed to do exactly that: assess the relevance of different sentences to the symptoms of depression as outlined in the BDI questionnaire. This report is all about how our team, Mason-NLP, participated in this subtask, which involved identifying sentences related to different depression symptoms. We used a deep learning approach that incorporated MentalBERT, RoBERTa, and LSTM. Despite our efforts, the evaluation results were lower than expected, underscoring the challenges inherent in ranking sentences from an extensive dataset about depression, which necessitates both appropriate methodological choices and significant computational resources. We anticipate that future iterations of this shared task will yield improved results as our understanding and techniques evolve.

\end{abstract}

\begin{keywords}
  depression detection \sep
  early risk prediction \sep
  mental health \sep
  natural language processing \sep
  information retrieval \sep
  CEUR-WS
\end{keywords}

\maketitle
\section{Introduction}

Depression is an increasingly prevalent mental health condition, particularly among teens and younger individuals \cite{thapar2012depression}. Each year, millions are affected by this debilitating condition, yet a substantial proportion do not seek medical attention, due to either lack of awareness or the stigma associated with mental health disorders \cite{andrade2014barriers}. This untreated condition can have profound personal and social implications, potentially leading to severe consequences such as substance abuse \cite{sinha2008chronic} or even suicide \cite{nock2010mental}. 

Language is more than a medium for communication; it's a reflection of various aspects of our lives. Language can unveil a multitude of insights about us, ranging from our age and gender to our upbringing and mental state \cite{pennebaker2003psychological}. Evidence suggests a correlation between language use and mental health \cite{tausczik2010psychological}, positioning natural language processing (NLP) as a potent tool for analyzing depression symptoms. Indeed, recent research indicates the efficacy of these techniques in detecting signs and symptoms of various mental health conditions \cite{de2013social}.

In today's digital age, a large proportion of people, especially youth and teens, actively use social media \cite{perrin2015social}. This widespread use, and the potential correlation between increased social media interaction and depression, opens up possibilities for mental health analysis \cite{lin2016association}. Social media serves as a platform for individuals to communicate, share thoughts and experiences, and engage in topic-specific channels like subreddits. Particularly, channels related to mental health and depression can provide valuable insights when analyzed with state-of-the-art NLP techniques \cite{de2013social, coppersmith2015adhd}.

Early detection of depression symptoms is essential since it allows for timely intervention and lowers the risk of serious effects \cite{cuijpers2016effective}. With their wealth of user-generated content, social media platforms can be a powerful tool for this early detection \cite{ de2013social, coppersmith2015adhd}. If it is possible to extract words from social media posts that are pertinent to depressive symptoms, preventative steps to treat probable mental health issues could be taken \cite{birnbaum2018digital}.

We give a thorough explanation of our participation in eRisk 2023 Task 1 in our publication. This project focuses on identifying early risks on the internet, and in particular, our task comprises ranking user data into sentences depending on how closely they correspond to symptoms of depression as identified by the Beck Depression Inventory (BDI) Questionnaire. Please see \cite{parapar2023erisk} for further details regarding the task and a detailed summary of the evaluations.
\section{Dataset}

Our system was trained utilizing two distinct datasets: 1) the official dataset for the eRisk 2023 Task 1, and 2) the "Depression: Reddit Dataset" from Kaggle\cite{kaggle}.

The official dataset for the eRisk 2023 Task 1, derived from previous eRisk data, was formatted in accordance with Text REtrieval Conference(TREC)\cite{trec} guidelines. This dataset consisted of a collection of user documents, with 3,107 unique users contributing to a total of 3,807,115 sentences. On average, each user had approximately 1,225 sentences associated with their profile.

However, inherent challenges with this data source existed due to its nature. Derived from informal social media posts, the sentences often lacked cohesiveness, contained spelling and grammatical errors, and frequently included internet shorthand. Additionally, the presence of links, images, references to other users, and a high volume of sentences irrelevant to the task at hand posed further complexities. Therefore, identifying pertinent sentences within this extensive corpus presented a significant challenge.

The second dataset, the "Depression: Reddit Dataset" from Kaggle, comprised raw data scraped from various mental health-related subreddits. Post-scraping, the data was cleaned and processed using different NLP techniques to prepare it for depression classification. This dataset contained 7,731 sentences, each associated with a binary label: '1' representing depression-related sentences and '0' denoting non-depressed sentences. The dataset was fairly balanced, containing 3,831 positive examples and 3,900 negative examples.
\section{Methodology and experimental setup }
Our methodology leveraged three models in a dual-stage approach, as described in Figure 1, aimed at accelerating system performance. We will first present a summary of each model before diving into the specifics of our technique.
\begin{figure}
  \centering
  \includegraphics[width=0.9\textwidth]{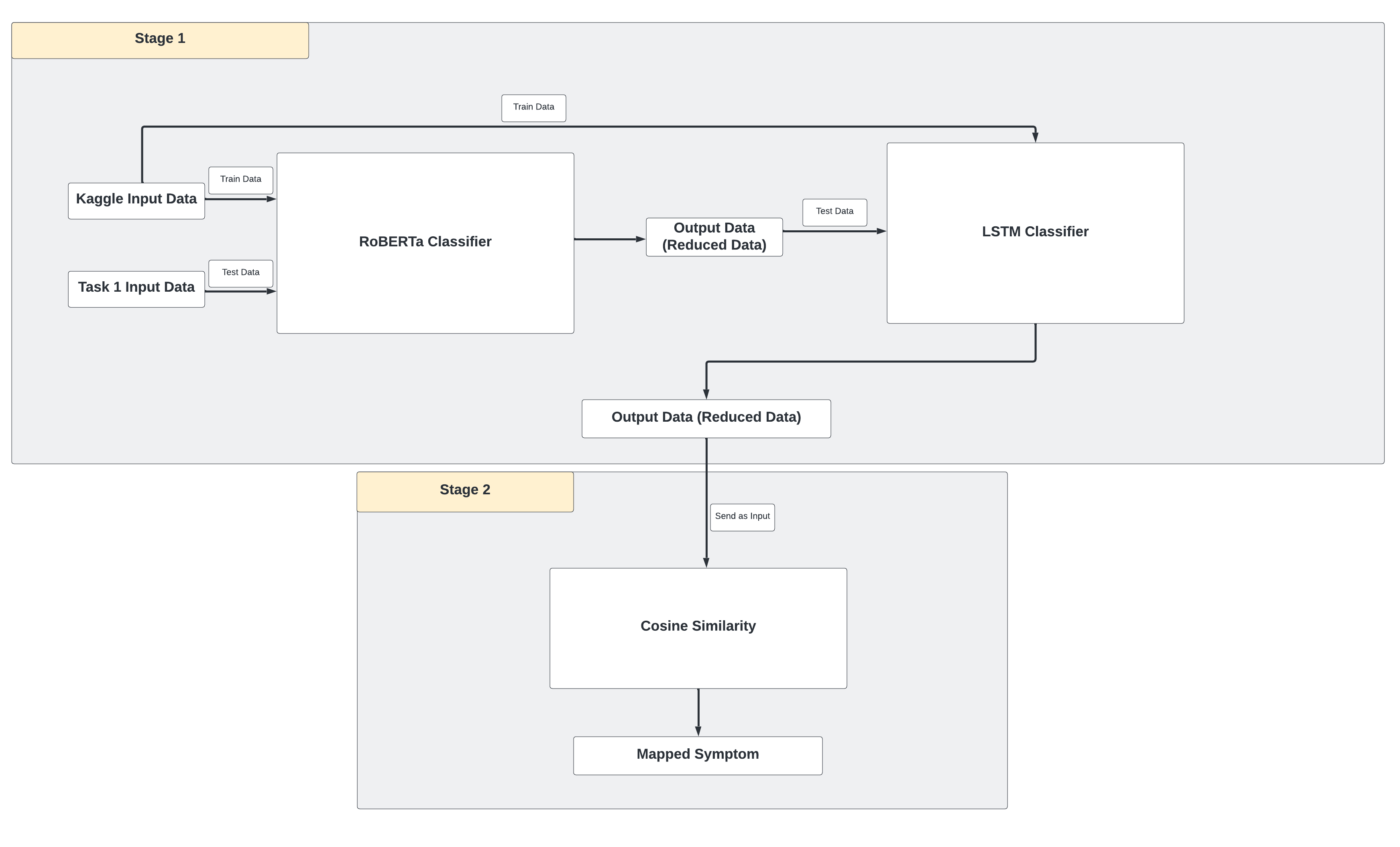}
  \caption{Methodology Flowchart}
  \label{fig:example}
\end{figure}

\paragraph{MentalBERT}
MentalBERT\cite{liu2019roberta} is a pre-trained language model for mental healthcare pre-trained on mental health data. Specifically, the pretraining data was collected from Reddit \cite{reddit}, especially from Mental Health related subreddits, like "r/depression","r/Anxiety", "r/bipolar" etc. The pretrained model was evaluated on various mental health detection tasks such as depression and stress detection. The empricial results show that pretraining with a mental health corpus improves classification performance. 
\paragraph{RoBERTa}
RoBERTa\cite{ji2021mentalbert}, short for Robustly Optimized BERT Pretraining Approach is a variant of BERT that enhances the pretraining process by adjusting key hyperparameters and training on a larger dataset. Unlike BERT, RoBERTa omits the next sentence prediction objective, focusing only on the masked language modeling task, and uses a dynamic masking pattern rather than a static one. Additionally, RoBERTa uses larger batch sizes during training. The changes incorporated in RoBERTa have led to superior performance over BERT on a variety of natural language understanding tasks, making it a powerful model for various applications in the field of natural language processing.
\paragraph{LSTM}
Long Short-Term Memory (LSTM)\cite{staudemeyer2019understanding} is a type of Recurrent Neural Network (RNN) used for processing and predicting time-series and sequential data. LSTMs manage long-term dependencies through a unique architecture involving a cell state and three gates: input, forget, and output gates. This architecture effectively mitigates the vanishing gradient problem seen in traditional RNNs, allowing LSTMs to learn and remember information over extended sequence lengths. This makes them ideal for tasks like handwriting or speech recognition and other sequence-to-sequence learning tasks.

The task at hand necessitated a bifurcated approach for optimal efficiency, with the primary objective of curtailing the data sample size. This reduction was principally motivated by the simultaneous application of multiple models in conjunction with the MentalBERT pre-trained model. These additional models were deployed during an exploratory phase, with the objective of identifying the most appropriate model for our system's architecture. Nevertheless, due to constraints in computational resources, executing multiple models on the comprehensive dataset, comprised of 3.8 million data points, resulted in considerable time consumption during both the training and testing phases. Consequently, to counteract this issue, a decision was made to eliminate sentences that were irrelevant to depressive symptoms. This strategy aimed to expedite the process of identifying a model that was best suited for our system.

The first phase of our task, therefore, concentrated on a standardized dataset specifically related to depression, sourced from the renowned data science platform, Kaggle. This depression-oriented dataset was employed to train a binary-model system designed to exclude sentences unrelated to depression.

The dual-model system comprised a transformer model, RoBERTa, as its initial component. After its training with the depression-specific dataset from Kaggle, RoBERTa demonstrated high accuracy of 92\%. on the validation data. This model was subsequently used to categorize sentences associated with depression, thereby reducing the data samples from 3,807,115 to a more manageable 830,151.

To further refine the sentence pool, the RoBERTa model's output was passed to the second component of our ensemble: a unidirectional, single-layer Long Short-Term Memory (LSTM) model. Also trained on the Kaggle dataset, this LSTM model contained 128 parameters and had a 20\% dropout rate, along with one fully connected layer leading to a sigmoid function. Trained over 20 epochs, the LSTM model achieved an impressive validation data accuracy of 97\%. After processing the RoBERTa model's reduced data, the LSTM model further diminished the sample size from 830,151 to 305,118, thus producing a more relevant and manageable dataset.

The second phase of our task, following the successful reduction of the initial dataset from 3,807,115 to 305,118 pertinent sentences, entailed a detailed comparison process. Each sentence from the reduced dataset was compared with a set of 21 distinct depression symptoms. Each symptom was articulated through four variations, yielding 84 unique sentences (21 symptoms * 4 variations) representing the depression symptom spectrum.

This comparison was conducted using cosine similarity, a metric capturing the cosine of the angle between two vectors—in our context, signifying the similarity between each sample sentence and the symptom sentence. We utilized a pre-trained model, MentalBERT, specifically chosen due to its relevance to our task. Trained on mental health-related datasets, MentalBERT is adept at interpreting and analyzing text in this specific domain.

For computation, sentences were tokenized, divided into constituent words or phrases, facilitating their input into the model. This tokenization was optimized through padding and CUDA \cite{cuda}, an application programming interface model created by NVIDIA\cite{nvidia}, resulting in accelerated processing through GPU power utilization. Each sentence, irrespective of length, was represented by an equal number of tokens. Following tokenization, embeddings for each token were computed. These embeddings transformed discrete tokens into continuous vectors, capturing semantic meaning and making them suitable for processing by machine learning models. Mean pooling was then applied to these embeddings, down-sampling the output. The mean-pooled sentence-embedded tokens were subsequently normalized, ensuring the vectors had a norm of one, ready to be input into the cosine similarity metric model.

Each sentence was compared with the 84 symptom sentences as it passed through the model, generating 84 cosine similarity scores for each sentence. The highest score among these represented the closest match, and the corresponding symptom was then mapped as the identified symptom for the given sample sentence. For each symptom, the similarity scores of the mapped sentences were arranged in descending order, retaining only the top 1,000 sentences for each symptom.

To sum up, the methodology developed in this research effectively amalgamates machine learning (ML) and natural language processing (NLP) techniques to handle an extensive dataset related to depression. Commencing with a sizeable, heterogeneous dataset, we applied ensemble learning models to pare down the sample to a manageable size. Subsequently, we leveraged sentence transformers and cosine similarity measures to assign each sentence to the most closely matching symptom of depression. This sophisticated two-stage approach, which combines several machine learning models with cosine similarity, furnished us with an effective and efficient method for processing large-scale depression-related data. Notably, the success of this approach underscores the potential of integrating multiple machine learning techniques to improve the processing and analysis of large and complex datasets in the field of mental health.

\section{Results and Discussion}
\begin{table}[]
\caption{Ranking-based evaluation:majority voting. Our results(Mason-NLP) compared with other teams' results. The best results are bolded. }
\begin{tabular}{llllll}
Team       & Run                                      & AP             & R-PREC         & P@10           & NDCG@1000      \\ \hline
Mason-NLP  & MentalBert                               & 0.035          & 0.072          & 0.286          & 0.117          \\ \hline
Formula-ML & SentenceTrainsformers\_0.25              & \textbf{0.319} & \textbf{0.375} & \textbf{0.861} & \textbf{0.596} \\
Formula-ML & SentenceTrainsformers\_0.1               & 0.308          & 0.359          & \textbf{0.861} & 0.584          \\
Formula-ML & result2                                  & 0.086          & 0.170          & 0.457          & 0.277          \\
Formula-ML & word2vec\_0.1                            & 0.092          & 0.176          & 0.5            & 0.285          \\
OBSER-MENH & salida-distilroberta-90-cos              & 0.294          & 0.359          & 0.814          & 0.578          \\
OBSER-MENH & salida-mpnet-90-cos                      & 0.265          & 0.333          & 0.805          & 0.550          \\
OBSER-MENH & salida-mpnet-21-cos                      & 0.120          & 0.207          & 0.471          & 0.365          \\
OBSER-MENH & salida-distilroberta-21-cos              & 0.158          & 0.249          & 0.543          & 0.418          \\
OBSER-MENH & salida-mini12-21-cos                     & 0.114          & 0.184          & 0.305          & 0.329          \\
uOttawa    & USESim                                   & 0.160          & 0.248          & 0.600          & 0.382          \\
uOttawa    & Glove100Sim                              & 0.017          & 0.052          & 0.195          & 0.105          \\
uOttawa    & RobertaSim                               & 0.033          & 0.080          & 0.329          & 0.150          \\
uOttawa    & GloveSim                                 & 0.011          & 0.038          & 0.162          & 0.075          \\
uOttawa    & BertSim                                  & 0.084          & 0.150          & 0.505          & 0.271          \\
BLUE       & SemSearchOnBDI2Queries                   & 0.104          & 0.126          & 0.781          & 0.211          \\
BLUE       & SemSearchOnGeneratedQueriesMentalRoberta & 0.029          & 0.063          & 0.367          & 0.105          \\
BLUE       & SemSearchOnBDI2QueriesMentalRoberta      & 0.027          & 0.044          & 0.386          & 0.089          \\
BLUE       & SemSearchOnGeneratedQueries              & 0.052          & 0.074          & 0.586          & 0.139          \\
BLUE       & SemSearchOnAllQueries                    & 0.065          & 0.086          & 0.629          & 0.160          \\
NailP      & T1\_M2                                   & 0.095          & 0.146          & 0.519          & 0.226          \\
NailP      & T1\_M4                                   & 0.095          & 0.146          & 0.519          & 0.221          \\
NailP      & T1\_M3                                   & 0.073          & 0.114          & 0.471          & 0.180          \\
NailP      & T1\_M5                                   & 0.089          & 0.140          & 0.486          & 0.223          \\
NailP      & T1\_M1                                   & 0.074          & 0.114          & 0.471          & 0.189          \\
RELAI      & bm25|mpnetbase                           & 0.048          & 0.081          & 0.538          & 0.140          \\
RELAI      & BM25                                     & 0.016          & 0.061          & 0.043          & 0.145          \\
RELAI      & bm25|mpnetbase\_simcse                   & 0.030          & 0.066          & 0.390          & 0.114          \\
RELAI      & bm25|mpnetqa\_simcse                     & 0.027          & 0.063          & 0.376          & 0.109          \\
RELAI      & bm25|mpnetqa                             & 0.038          & 0.075          & 0.438          & 0.126          \\
UNSL       & Prompting-Classifier                     & 0.036          & 0.090          & 0.229          & 0.180          \\
UNSL       & Similarity-AVG                           & 0.001          & 0.008          & 0.010          & 0.016          \\
UNSL       & Similarity-MAX                           & 0.001          & 0.011          & 0.019          & 0.019          \\
UMU        & LexiconMultilingualSentenceTransformer   & 0.073          & 0.140          & 0.495          & 0.222          \\
UMU        & LexiconSentenceTransformer               & 0.054          & 0.122          & 0.362          & 0.191          \\
GMU        & FAST-DCMN-COS-INJECT                     & 0.001          & 0.002          & 0.014          & 0.004          \\
GMU        & FAST-DCMN-COS-INJECT\_FULL               & 0.001          & 0.003          & 0.014          & 0.005         
\end{tabular}
\label{tab:my-table}
\end{table}
\begin{table}[]
\caption{Ranking-based evaluation: unanimity. Our results(Mason-NLP) compared with other teams’ results.The best results are bolded}
\begin{tabular}{llllll}
Team       & Run                                      & AP             & R-PREC         & P@10           & NDCG@1000      \\ \hline
Mason-NLP  & MentalBert                               & 0.024          & 0.054          & 0.190          & 0.099          \\ \hline
Formula-ML & SentenceTransformers\_0.25               & 0.268          & \textbf{0.360} & \textbf{0.709} & \textbf{0.615} \\
Formula-ML & SentenceTransformers\_0.1                & \textbf{0.293} & 0.350          & 0.685          & 0.611          \\
Formula-ML & result2                                  & 0.079          & 0.155          & 0.357          & 0.290          \\
Formula-ML & word2vec\_0.1                            & 0.085          & 0.163          & 0.357          & 0.299          \\
OBSER-MENH & salida-distilroberta-90-cos              & 0.281          & 0.344          & 0.652          & 0.604          \\
OBSER-MENH & salida-mpnet-90-cos                      & 0.252          & 0.337          & 0.643          & 0.575          \\
OBSER-MENH & salida-distilroberta-21-cos              & 0.135          & 0.216          & 0.390          & 0.413          \\
OBSER-MENH & salida-mini12-21-cos                     & 0.099          & 0.165          & 0.214          & 0.329          \\
OBSER-MENH & salida-mpnet-21-cos                      & 0.101          & 0.189          & 0.319          & 0.366          \\
uOttawa    & USESim                                   & 0.139          & 0.232          & 0.438          & 0.380          \\
uOttawa    & GloveSim                                 & 0.008          & 0.028          & 0.110          & 0.063          \\
uOttawa    & Glove100Sim                              & 0.011          & 0.042          & 0.110          & 0.092          \\
uOttawa    & RobertaSim                               & 0.025          & 0.068          & 0.190          & 0.140          \\
uOttawa    & BertSim                                  & 0.070          & 0.130          & 0.357          & 0.260          \\
BLUE       & SemSearchOnBDI2Queries                   & 0.129          & 0.167          & 0.643          & 0.260          \\
BLUE       & SemSearchOnAllQueries                    & 0.067          & 0.105          & 0.452          & 0.177          \\
BLUE       & SemSearchOnGeneratedQueriesMentalRoberta & 0.018          & 0.059          & 0.186          & 0.085          \\
BLUE       & SemSearchOnGeneratedQueries              & 0.052          & 0.088          & 0.381          & 0.147          \\
BLUE       & SemSearchOnBDI2QueriesMentalRoberta      & 0.032          & 0.058          & 0.300          & 0.104          \\
NailP      & T1\_M2                                   & 0.090          & 0.143          & 0.410          & 0.229          \\
NailP      & T1\_M4                                   & 0.090          & 0.143          & 0.410          & 0.224          \\
NailP      & T1\_M5                                   & 0.083          & 0.139          & 0.338          & 0.222          \\
NailP      & T1\_M1                                   & 0.073          & 0.114          & 0.343          & 0.192          \\
NailP      & T1\_M3                                   & 0.073          & 0.114          & 0.343          & 0.181          \\
UMU        & LexiconSentenceTransformer               & 0.044          & 0.110          & 0.210          & 0.175          \\
UMU        & LexiconMultilingualSentenceTransformer   & 0.059          & 0.125          & 0.333          & 0.209          \\
RELAI      & BM25                                     & 0.012          & 0.036          & 0.019          & 0.135          \\
RELAI      & bm25|mpnetbase\_simcse                   & 0.026          & 0.059          & 0.243          & 0.103          \\
RELAI      & bm25|mpnetqa\_simcse                     & 0.023          & 0.052          & 0.262          & 0.097          \\
RELAI      & bm25|mpnetqa                             & 0.030          & 0.065          & 0.290          & 0.109          \\
RELAI      & bm25|mpnetbase                           & 0.039          & 0.069          & 0.343          & 0.124          \\
UNSL       & Similarity-MAX                           & 0.001          & 0.006          & 0.010          & 0.012          \\
UNSL       & Prompting-Classifier                     & 0.020          & 0.063          & 0.090          & 0.157          \\
UNSL       & Similarity-AVG                           & 0.000          & 0.005          & 0.005          & 0.011          \\
GMU        & FAST-DCMN-COS-INJECT\_FULL               & 0.001          & 0.003          & 0.014          & 0.006          \\
GMU        & FAST-DCMN-COS-INJECT                     & 0.001          & 0.002          & 0.010          & 0.003         
\end{tabular}
\label{tab:my-table1}
\end{table}

In this section, we present the results of our participation in the eRisk 2023 Task 1 as well as discuss potential reasons for the performance of our model. Given the task's nature as an Information Retrieval challenge, there was no designated test dataset. Instead, three independent assessors, a trained psychologist and two experts in early risk technologies, evaluated the results. The assessors assigned relevance scores based on two criteria: '1' for sentences relevant to a BDI symptom, and '0' for sentences not relevant or not conveying any information about a BDI symptom.

Various standard performance metrics were reported, including Average Precision (AP), R-Precision, Precision at 10, and Normalized Discounted Cumulative Gain (NDCG) at 1000, based on two aggregation criteria: unanimity and majority voting.
The results are reported in Table 1 and Table 2. 

Our system, which employed MentalBERT, obtained scores of 0.035, 0.072, 0.286, and 0.117 in AP, R-Precision, Precision at 10, and NDCG at 1000 respectively under majority voting. Under unanimity aggregation, our scores were lower: 0.024, 0.054, 0.190, and 0.099 respectively. The top-performing systems in the task significantly outperformed our model.

While our expectations were higher given MentalBERT's domain-specific training, we can identify several potential factors that might have influenced our system's performance:

Training Data: While MentalBERT is pre-trained on a mental health data corpus, it may not resonate perfectly with the nuances of the BDI Questionnaire, which focuses on subjective experiences of depression symptoms. That's why it might have struggled to accurately assess relevance.

Evaluation Criteria: The utilized metrics (AP, R-Precision, Precision at 10, and NDCG at 1000) favor systems retrieving relevant sentences ranked high. If our system retrieved relevant sentences ranked lower, it would negatively affect these scores.

Model Tuning and Parameters: Models often require task-specific tuning and adjustment. Fine-tuning MentalBERT or adjusting its parameters for this task might have enhanced performance.

Unanimity vs. Majority Voting: Our system's better performance in majority voting suggests it might be detecting more ambiguous or less agreed-upon symptom instances, leading to lower performance when higher agreement levels are required.

These insights provide useful direction for future work and refinement of our approach.

\section{Conclusion and Future Work}
This paper has documented the approach and outcomes of MASON-NLP's submission for the eRisk 2023 Task 1, employing a two-step methodology utilizing ROBERTA, MentalBERT and LSTM to retrieve sentences pertinent to BDI symptoms from a sizeable Reddit dataset.

In conclusion, the demonstrated performance of our methodology, albeit reasonable, did not match up to the top-tier teams, underlining room for improvement. The Mental BERT model's capacity to resonate with the task requirements could be optimized through fine-tuning using task-specific data, which emerges as an immediate future direction.

Understanding the nuances of evaluation metrics and aligning our strategies to maximize these metrics could lead to significant gains in performance. A closer inspection of our model's performance under unanimity aggregation criteria, followed by strategies to improve in this area, also warrants attention.

In essence, our future work will focus on learning from this experience and making informed, strategic modifications to our approach to enhance its effectiveness in identifying BDI symptoms from social media text data. This study's insights underscore the challenges and possibilities inherent in leveraging AI for mental health, underlining its significance in the ongoing dialogue around technology's role in healthcare.

\bibliography{sample-ceur}




\end{document}